\documentclass{article}

\usepackage[preprint,nonatbib]{nips_2018}

\usepackage[utf8]{inputenc} 
\usepackage[T1]{fontenc}    
\usepackage{hyperref}       
\usepackage{url}            
\usepackage{booktabs}       
\usepackage{amsfonts}       
\usepackage{nicefrac}       
\usepackage{microtype}      

\usepackage{floatrow}
\usepackage{color}
\usepackage{amsmath, amsthm, amssymb}
\usepackage{bm}
\usepackage{graphicx}
\usepackage{multirow}
\usepackage{tabularx}
\usepackage{booktabs}
\usepackage[ruled,vlined]{algorithm2e}
\usepackage{algorithmic}
\usepackage{float}
\usepackage{wrapfig}
\usepackage{subfigure}

\newcommand{\bdot}[0]{\boldsymbol{\cdot}}

\newcommand{\softplus}[0]{+}
\newcommand{\ep}[2]{\mathbb{E}_{#1}\sbb{#2}}

\newcommand{\KL}[2]{\mathrm{KL}\rbb{#1||\;#2}}
\newcommand{\entropy}[1]{\mathrm{H}\sbb{#1}}

\newcommand{\bx}[0]{\mathbf{x}}
\newcommand{\ba}[0]{\mathbf{a}}
\newcommand{\bw}[0]{\mathbf{w}}

\newcommand{\bwx}[0]{\bw_x}
\newcommand{\bwa}[0]{\bw_a}

\newcommand{\sx}[0]{\tilde{\bx}}
\newcommand{\sa}[0]{\tilde{\ba}}

\newcommand{\rbb}[1]{\left( #1 \right)}
\newcommand{\cbb}[1]{\left\lbrace #1 \right\rbrace}
\newcommand{\sbb}[1]{\left[ #1 \right]}

\newcommand{\demo}[0]{\text{demo}}
\newcommand{\train}[0]{\text{train}}
\newcommand{\loss}[0]{\mathcal{L}^\mathrm{ML}}
\newcommand{\losskl}[0]{\mathcal{L}^\mathrm{KL}}

\newcommand{\cprob}[2]{p\rbb{#1 \;|\; #2}}

\newcommand{\csamplex}[2]{\pi_x\rbb{\;#1\;|\;#2}}
\newcommand{\csamplea}[2]{\pi_a\rbb{\;#1\;|\;#2}}
\newcommand{\samplex}[0]{\pi_x}
\newcommand{\samplea}[0]{\pi_a}

\DeclareMathOperator*{\argmin}{argmin}

\newcommand{\insertfigure}[3]{
\begin{figure}[ht]
\centering
\includegraphics[width=#2\textwidth]{fig/#1}
\caption{#3}
\label{fig:#1}
\end{figure}
}

\newcommand{\insertwrapfigure}[4]{
\begin{wrapfigure}{r}{#3\textwidth}
\centering
\includegraphics[width=#2\textwidth]{fig/#1}
\caption{#4}
\label{fig:#1}
\end{wrapfigure}
}

\newcommand{\insertsidefigure}[4]{
\begin{figure}[ht]
\floatbox[{\capbeside\thisfloatsetup{capbesideposition={right,center},capbesidewidth=#3\textwidth}}]{figure}[\FBwidth]
{\caption{#4}\label{fig:#1}}
{\includegraphics[width=#2\textwidth]{fig/#1}}
\end{figure}
}

\newcommand{\inserttable}[4]{
\begin{table}[H]
\centering
\begin{tabular}{#2}
#3
\end{tabular}
\label{table:#1}
\end{table}
}

\title{Concept Learning with Energy-Based Models}

\author{
  Igor Mordatch \\
  OpenAI\\
  San Francisco, CA \\
  \texttt{mordatch@openai.com} \\
}

\begin{document}

\maketitle


\begin{abstract}
Many hallmarks of human intelligence, such as generalizing from limited experience, abstract reasoning and planning, analogical reasoning, creative problem solving, and capacity for language require the ability to consolidate experience into \emph{concepts}, which act as basic building blocks of understanding and reasoning.
We present a framework that defines a concept by an energy function over events in the environment, as well as an attention mask over entities participating in the event. Given few demonstration events, our method uses inference-time optimization procedure to generate events involving similar concepts or identify entities involved in the concept.
We evaluate our framework on learning visual, quantitative, relational, temporal concepts from demonstration events in an unsupervised manner. Our approach is able to successfully generate and identify concepts in a few-shot setting and resulting learned concepts can be reused across environments.
Example videos of our results are available at \textbf{\href{https://sites.google.com/site/energyconceptmodels/}{sites.google.com/site/energyconceptmodels}}
\end{abstract}
\section{Introduction}
Many hallmarks of human intelligence, such as generalizing from limited experience, abstract reasoning and planning, analogical reasoning, creative problem solving, and capacity for language and explanation are still lacking in the artificial intelligent agents. We, as others \cite{rosch76basicobjects,lakoff1980metaphorical,lake2016building} believe what enables these abilities is the capacity to consolidate experience into \emph{concepts}, which act as basic building blocks of understanding and reasoning.

Examples of concepts include visual (\emph{"red"} or \emph{"square"}), spatial (\emph{"inside"}, \emph{"on top of"}), temporal (\emph{"slow"}, \emph{"after"}), social (\emph{"aggressive"}, \emph{"helpful"}) among many others \cite{lakoff1980metaphorical}. These concepts can be either identified or generated - one can not only find a square in the scene, but also create a square, either physical or imaginary. Importantly, humans also have a largely unique ability to combine concepts compositionally (\emph{"red square"}) and recursively (\emph{"move inside moving square"}) - abilities reflected in the human language. This allows expressing an exponentially large number of concepts, and acquisition of new concepts in terms of others. We believe the operations of identification, generation, composition over concepts are the tools with which intelligent agents can understand and communicate existing experiences and reason about new ones.

Crucially, these operations must be performed on the fly throughout the agent's execution, rather than merely being a static product of an offline training process. Execution-time optimization, as in recent work on meta-learning \cite{finn2017maml} plays a key role in this. We pose the problem of parsing experiences into an arrangement of concepts as well as the problems of identifying and generating concepts as optimizations performed during execution lifetime of the agent. The meta-level training is performed by taking into account such processes in the inner level.

Specifically, a concept in our work is defined by an energy function taking as input an event configuration (represented as trajectories of entities in the current work), as well as an attention mask over entities in the event. Zero-energy event and attention configurations imply that event entities selected by the attention mask satisfy the concept. Compositions of concepts can then be created by simply summing energies of constituent concepts. Given a particular event, optimization can be used to identify entities belonging to a concept by solving for attention mask that leads to zero-energy configuration. Similarly, an example of a concept can be generated by optimizing for a zero-energy event configuration. See Figure \ref{fig:example} for examples of these two processes.

The energy function defines a family of concepts, from which a particular concept is selected with a specific concept code. Encoding of event and attention configurations can be achieved by execution-time optimization over concept codes. Once an event is encoded, the resulting concept code structure can be used to re-enact the event under different initial configurations (task of imitation learning), recognize similar events, or concisely communicate the nature of the event. We believe there is a strong link between concept codes and language, but leave it unexplored in this work.

At the meta level, the energy function is the only entity that needs to be learned. This is different from generative model or inverse reinforcement learning approaches, which typically also learn an explicit generator/policy function, whereas we define it implicitly via optimization. Our advantage as that the learned energy function can be reused in other domains, for example using a robot platform to re-enact concepts in the physical world. Such transfer is not possible with an explicit generation/policy function, as it is domain-specific.
\insertfigure{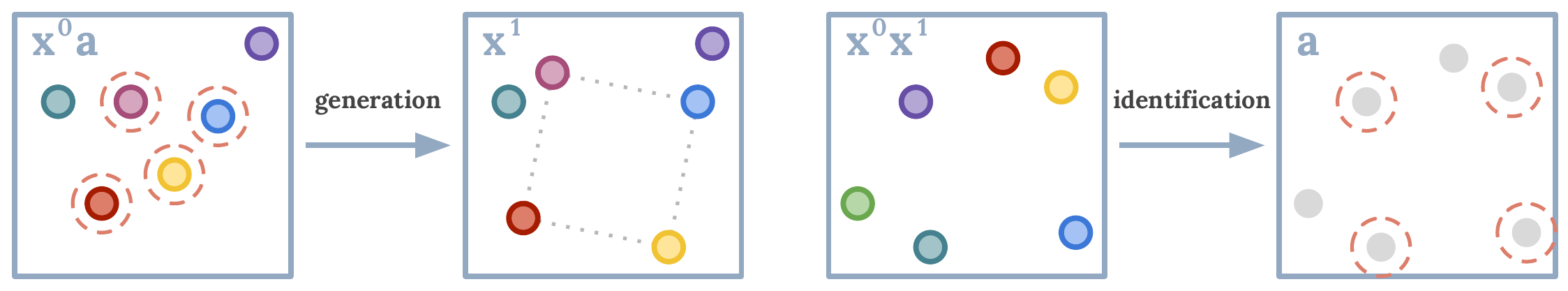}{1.0}{Examples of generation and identification processes for a \emph{"square"} concept. a) Given initial state $\bx^0$ and attention mask $\ba$, square consisting of entities in $\ba$ is formed via optimization over $\bx^1$. b) Given states $\bx$, entities comprising a square are found by optimization over attention mask $\ba$.}

\section{Related Work}
We draw upon several areas for inspiration in our work, including energy-based models, concept learning, inverse reinforcement learning and meta-learning.

Energy-based modelling approaches have a long history in machine learning, commonly used for density modeling \cite{dayan1995helmholtz,hinton2006training,salakhutdinov2009deep,friston2010free}. These approaches typically aim to learn a function that assigns low energy values to inputs in the data distribution and high energy values to other inputs. The resulting models can then be used to either discriminate whether or not a query input comes from the data distribution, or to generate new samples from the data distribution. One common choice for sampling procedure is Markov Chain Monte Carlo (MCMC), however it suffers from slow mixing and often requires many iterative steps to generate samples \cite{salakhutdinov2009deep}. Another choice is to train a separate network to generate samples \cite{kim2016deep}. Generative Adversarial Networks \cite{goodfellow2014generative} can be thought of as instances of this approach \cite{finn2016connection}. The key difficulty in training energy-based models lies in estimating their partition function, with many approaches relying on the sampling procedure to estimate it or further approximations \cite{hinton2006training}. Our approach avoids both the slow mixing of MCMC and the need to train a separate network. We use sampling procedure based on gradient of the energy (which mixes much faster than gradient-free MCMC), while training the energy function to have a gradient field that produces good samples.

The problem of learning and reasoning over concepts or other abstract representations has long been of interest in machine learning (see \cite{lake2016building,bengio2013representation} for review). Approaches based on Bayesian reasoning have notably been applied for numerical concepts \cite{tenenbaum1999bayesian}. A recent framework of \cite{higgins2017scan} focuses on visual concepts such as color and shape, where concepts are defined in terms of distributions over latent variables produced by a variational autoencoder. Instead of focusing solely on visual concepts from pixel input, our work explores learning of concepts that involve complex interaction of multiple entities.

Our method aims to learn concepts from demonstration events in the environment. A similar problem is tackled by inverse reinforcement learning (IRL) approaches, which aim to infer an underlying cost function that gives rise to demonstrated events. Our method's concept energy functions are analogous to the cost or negative of value functions recovered by IRL approaches. Under this view, multiple concepts can easily be composed simply by summing their energy functions. Concepts are then enacted in our approach via energy minimization, mirroring the application a forward reinforcement learning step in IRL methods. Max entropy \cite{ziebart2008maximum} is a common IRL formulation, and our method closely resembles recent instantiations of it \cite{finn2016guided,dvijotham2010inverse}.

Our method relies on performing inference-time optimization processes for concept generation and identification, as well as for determining which concepts are involved in an event. The training is performed by taking behavior of these inner optimization processes into account, similar to the meta-learning formulations of \cite{finn2017maml}. Relatedly, iterative processes have been explored in the context of control \cite{tamar2016value,silver2016predictron,weber2017imagination,hamrick2017metacontrol,silver2017mastering} and image generation \cite{gregor2015draw}.





\section{Energy-Based Concept Models}
\label{sec:model}
Concepts operate over events, which in this work is a trajectory of $T$ states $\bx = \left[ \bx^0, ..., \bx^T \right]$. Each state contains a collection of $N$ entities $\bx^t = \left[ \bx_0,..., \bx_N \right]$ and each entity $\bx^t_i$ can contain information such as position and color of the entity. Considering entire trajectories and entities allows us to model temporal or relational concepts, unlike work that focuses on visual concepts \cite{higgins2017scan}. Attention over entities in the event is specified by a mask $\ba \in \mathbb{R}^N$ over each of the entities.

Existence of a particular concept is given by energy function $E(\bx,\ba,\bw) \in \mathbb{R}^+$, where parameter vector $\bw$ specifies a particular concept from a family. The interpretation of $\bw$ is similar to that of a code in an autoencoder. $E(\bx,\ba,\bw) = 0$ when state trajectory $\bx$ under attention mask $\ba$ over entities satisfies the concept $\bw$. Otherwise, $E(\bx,\ba,\bw) > 0$. The conditional probabilities of a particular event configuration belonging to a concept and a particular attention mask identifying a concept are given by the Boltzmann distributions:
\begin{align}
\label{eq:probs}
p(\bx| \ba, \bw) \propto \exp\cbb{-E(\bx, \ba, \bw)} && p(\ba| \bx, \bw) \propto \exp\cbb{-E(\bx, \ba, \bw)}
\end{align}
Given concept code $\bw$, the energy function can be used for both generation and identification of a concept implicitly via optimization (see Figure \ref{fig:example}):
\begin{align}
\label{eq:argmin}
\bx(\ba) = \argmin_\bx E(\bx,\ba,\bw) && \ba(\bx) = \argmin_\ba E(\bx,\ba,\bw)
\end{align}
Samples from distributions in (\ref{eq:probs}) can be generated via stochastic gradient Langevin dynamics, effectively performing stochastic minimization in (\ref{eq:argmin}):
\begin{align}
\tilde{\bx} \sim \csamplex{\bdot}{\ba, \bw} &= \bx^K, \;\; \bx^k = \bx^{k-1} + \frac{\alpha}{2} \nabla_\bx E(\bx, \ba, \bw) + \omega^k \nonumber \\
\label{eq:sample}
\tilde{\ba} \sim \csamplea{\bdot}{\bx, \bw} &= \ba^K, \;\; \ba^k = \ba^{k-1} + \frac{\alpha}{2} \nabla_\ba E(\bx, \ba, \bw) + \omega^k, \; \; \; \omega^k \sim \mathcal{N}(0,\alpha)
\end{align}
This stochastic optimization procedure is performed during execution time of the algorithm and is reminiscent of the Monte Carlo sampling procedures in prior work on energy-based models \cite{hinton2006training,salakhutdinov2009deep,friston2010free}. The procedure differs from approaches that use explicit generator functions \cite{dayan1995helmholtz,kingma2013auto,kim2016deep} or explicit attention mechanisms, such as dot product attention \cite{olah2016attention}.

It is shown in \cite{welling2011bayesian} that $\tilde{\bx}$ and $\tilde{\ba}$ will approach samples from posterior distributions $p$ as $K \mapsto \infty$ and $\alpha \mapsto 0$. However in practice it is only possible to execute the dynamics for a finite number of steps (we use $K=10$ in all our experiments). This truncated procedure results in samples drawn from a biased distribution, which we call $\pi$ and which may not be equal to $p$. Similar issues of slow mixing are also present in prior work, which typically uses non-differentiable sampling procedures. In our case, the sampling procedure in equation (\ref{eq:sample}) can be differentiated and can be trained to produce samples close to true distribution $p$.

There are many possible choices for the energy function as long as it is non-negative. The specific form we use in this work is based on relation network architecture \cite{santoro2017simple} for its ability to easily capture interactions between pairs of entities
\begin{align}
\label{eq:energyfunc}
E_\theta(\bx,\ba,\bw) &= f_\theta(\sum_{t,i,j} \sigma(\ba_i) \sigma(\ba_j) \cdot g_\theta(\bx^t_i, \bx^t_j, \; \bw), \; \bw)^2
\end{align}
Where $f$ and $g$ are multi-layer neural networks that each take concept code as part of their input. $\sigma$ is the sigmoid function and is used to gate the entity pairs by their attention masks.

\section{Learning Concepts from Events}
\label{sec:learning}
To learn concepts from experience grounded in events, we pose a few-shot prediction task. Given a few demonstration examples $X^\demo$ containing tuples $\rbb{\bx, \ba}$ and initial state $\bx^0$ for a new event in $X^\train$, the task is to predict attention $\ba$ and the future state trajectory $\bx^{1:T}$ of the new event. The new event may contain a different configuration or number of entities, so it is not possible to directly transfer attention mask, for instance. To simplify notation, we consider prediction of only one future state $\bx^1$, although predicting more states is straightforward. The procedure is depicted in Figure \ref{fig:learning}.
\insertfigure{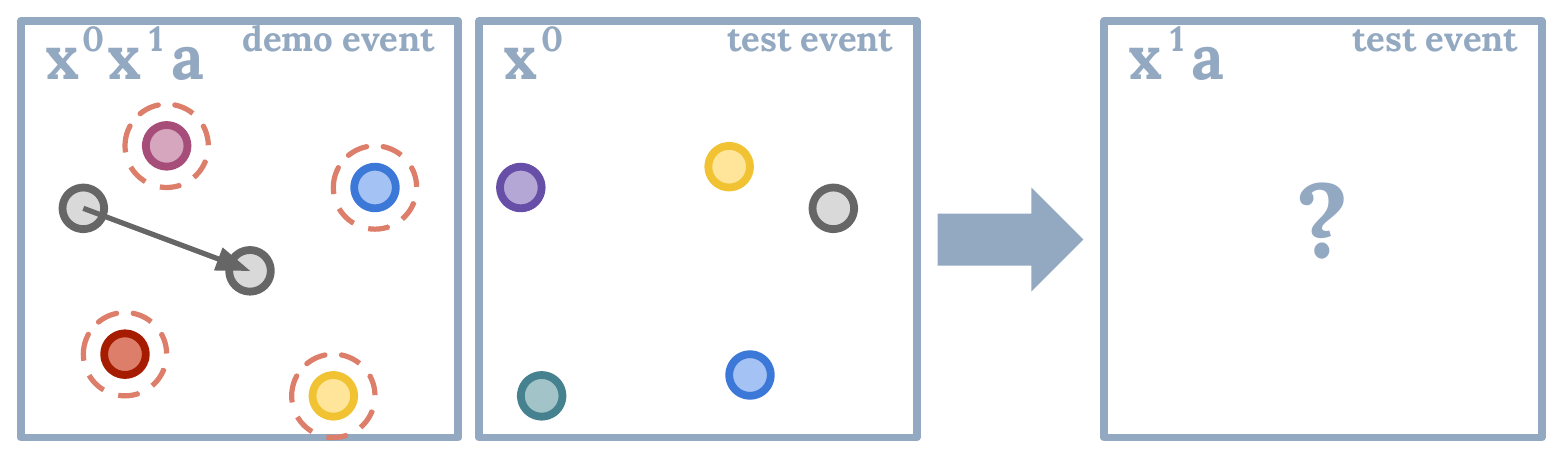}{0.65}{Example of a few-shot prediction task we use to learn concept energy functions.}


We follow the maximum entropy inverse reinforcement learning formulation \cite{ziebart2008maximum} and assume demonstrations are samples from the distributions given by the energy function $E$. Given an inferred concept code $\bw$ (details discussed below), finding energy function parameters $\theta$ is posed as as maximum likelihood estimation problem over future state and attention given initial state. The resulting loss for a particular dataset $X$ is
\begin{align}
\label{eq:loss_ml}
\loss_p(X,\bw) = \ep{\rbb{\bx, \ba} \sim X}{ -\log \cprob{\bx^1,\ba}{\bx^0,\bw} }
\end{align}

Where the joint probability can be decomposed in terms of probabilities in (\ref{eq:probs}) as
\begin{align}
\label{eq:joint}
\log\cprob{\bx^1,\ba}{\bx^0,\bw} = \log\cprob{\bx^1}{\ba,\bwx} + \log\cprob{\ba}{\bx^0,\bwa}, \; \; \; \bw = \sbb{\bwx, \bwa}
\end{align}

We use two concept codes, $\bwx$ and $\bwa$ to specify the joint probability. The interpretation is that $\bwx$ specifies the concept of the action that happens in the event (i.e. \emph{"be in center of"}) while $\bwa$ specifies the argument the action happens over (i.e. \emph{"square"}). This is a concept structure or syntax that describes the event. The concept codes are interchangeable and same concept code can be used either as action or as an argument because the energy function defining the concept can either be used for generation or identification. This importantly allows concepts to be understood from their usage under multiple contexts.


Conditioned on the two codes concatenated as $\bw$, the two log-likelihood terms in (\ref{eq:joint}) can be approximated as (see Appendix for the derivation)
\begin{align}
\label{eq:contrastive}
\log\cprob{\bx^1}{\ba,\bwx} \approx& -\sbb{E(\bx^1, \ba, \bwx) - E(\sx, \ba, \bwx)}_\softplus && \sx \sim \csamplex{\bdot}{\ba,\bwx} \nonumber \\
\log\cprob{\ba}{\bx^0,\bwa} \approx& -\sbb{E(\bx^0, \ba, \bwa) - E(\bx^0, \sa, \bwa)}_\softplus && \sa \sim \csamplea{\bdot}{\bx^0,\bwa}
\end{align}
Where $\sbb{\cdot}_\softplus = \log(1 + \exp(\cdot))$ is the softplus operator. This form is similar to contrastive divergence \cite{hinton2006training} and structured SVM forms \cite{belanger2017end} and is a special case of guided cost learning formulation \cite{finn2016guided}. The approximation comes from sample-based estimates of the partition functions for $p(\bx)$ and $p(\ba)$.

The above equations make use of truncated and biased gradient-based sampling distributions $\samplex$ and $\samplea$ in (\ref{eq:sample}) to estimate the respective partition functions. Following \cite{finn2016guided}, the approximation error in these estimates is minimal when KL divergence between biased distribution $\pi$ and true distribution $\exp\cbb{-E}/Z$ is minimized:
\begin{align*}
\label{eq:loss_kl}
\losskl_\pi(X,\bw) &= \KL{\pi_x}{p_x} + \KL{\pi_a}{p_a} \\ &= \ep{\rbb{\bx, \ba} \sim X}{ E(\sx, \ba, \bwx) + E(\bx^0, \sa, \bwa)} + \entropy{\samplex} + \entropy{\samplea} \\
&\sx \sim \csamplex{\bdot}{\ba,\bwx}, \;\; \sa \sim \csamplea{\bdot}{\bx^0,\bwa} \nonumber
\end{align*}
The above equation intuitively encourages sampling distributions $\pi$ to generate samples from low-energy regions.

\paragraph{Execution-Time Inference of Concepts}
Given a set of example events $X$, the concept codes can be inferred at execution-time via finding codes $\bw$ that minimize $\loss$ and $\losskl$.  Similar to \cite{grant2017concept}, in this work we only consider positive examples when adapting $\bw$ and ignore the effect that changing $\bw$ has on the sampling distribution $\pi$. The result is simply minimizing the energy functions wrt $\bw$ over the concept example events
\begin{align}
\bw^*_\theta(X) = \argmin_\bw \ep{\rbb{\bx, \ba} \sim X}{ E_\theta(\bx^1, \ba, \bwx) + E_\theta(\bx^0, \ba, \bwa) }
\end{align}
This minimization is similar to execution-time parameter adaptation and the inner update of meta-learning approaches \cite{finn2017maml}. We perform the optimization with stochastic gradient updates similar to equation (\ref{eq:sample}). This approach again implicitly infers codes at execution time via meta-learning using only the energy model as opposed to incorporating additional explicit inference networks.

\paragraph{Meta-Level Parameter Optimization}
We seek probability density functions $p$ that maximize the likelihood of training data $X$ via $\loss_p$ and simultaneously we seek sampling distributions $\pi$ that generate samples from $p$ via $\losskl_\pi$. In inverse reinforcement learning setting of \cite{finn2016guided} and \cite{fu2017learning}, these two objectives correspond to cost and policy function optimization are treated as separate alternating optimization problems because they operate over two different functions. However, in our case both $p$ and $\pi$ are implicitly a functions of the energy model and its parameters $\theta$, a dependence which we denote as $p(\theta)$ and $\pi(\theta)$. Consequently we can pose the problem as a single joint optimization
\begin{align}
\min_\theta \loss_{p(\theta)}(X^\train, \bw^*_\theta(X^\demo)) + \losskl_{\pi(\theta)}(X^\train, \bw^*_\theta(X^\demo))
\end{align}
We solve the above optimization problem via end-to-end backpropagation, differentiation through gradient-based sampling procedures. See Figure \ref{fig:overview} for an overview of our procedure and appendix for a detailed algorithm description.
\insertfigure{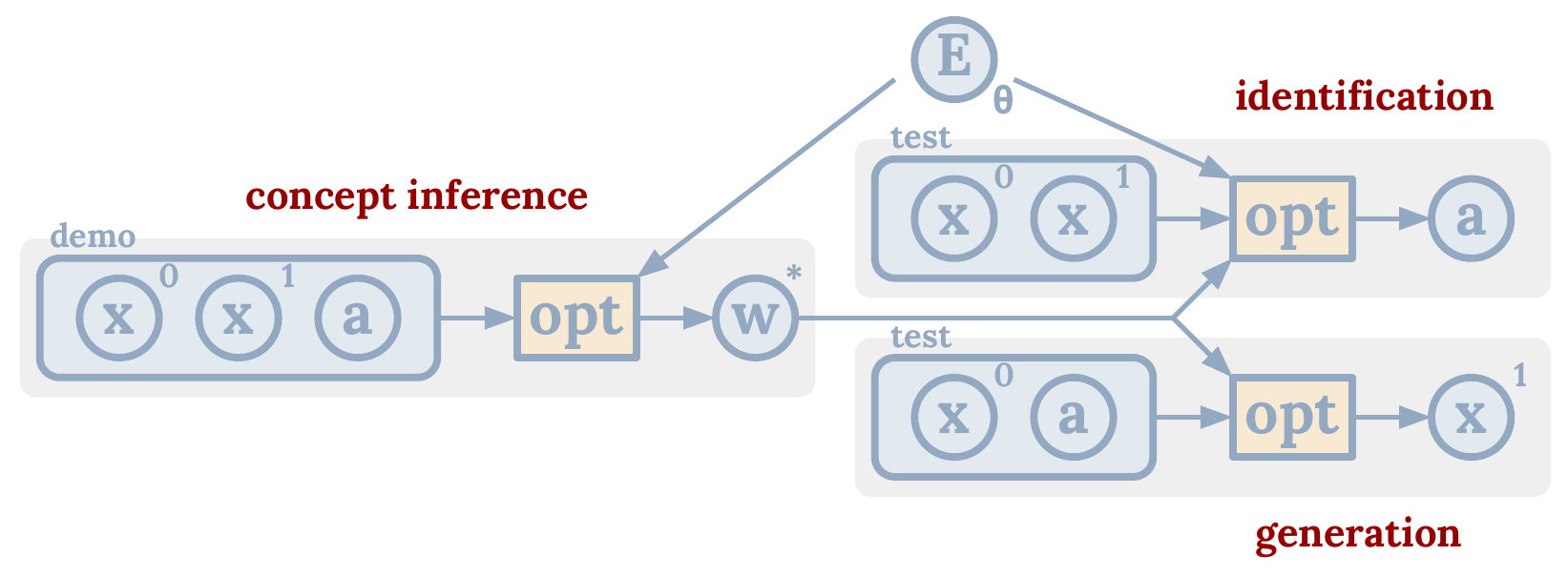}{0.75}{Execution-time inference in our method and the resulting optimization problems.}

\section{Experiments}
\label{sec:experiments}
The main purpose of our experiments is to investigate 1) whether our single model is able to learn understanding of wide variety of concepts under multiple contexts, 2) the utility of iterative optimization-based inference processes, and 3) ability to reuse learned concepts on different environments and actuation platforms.

\subsection{Evaluation Environment and Tasks}
We wish to evaluate understanding of concepts under multiple contexts - generation and identification. To the best of our knowledge we are not aware of any existing datasets or environments that simultaneously test both contexts for a wide range of concepts. For example, \cite{johnson2017clevr} tests understanding via question answering, while \cite{higgins2017scan} focuses on visual concepts. Thus to evaluate our method, we introduce a new simulated environment and tasks which extend the work in \cite{lowe2017multi}. The environment is a two-dimensional scene consisting of a varying collection of entities, each processing position, color, and shape properties. We wanted environment and tasks to be simple enough to facilitate ease of analysis, yet complex enough to lead to formation of a variety of concepts. In this work we focus on position-based environment representation, but a pixel-based representation and generation would be an exciting avenue for future work.

The task in this environment is, given $N$ demonstration events that involve (we use $N=5$) that involve identical attention and state changes under different situations, perform analogous behavior under $N$ novel test situations (by attending to analogous entities and performing analogous state changes). Such behavior is not unique and these may be multiple possible solutions. Because our energy model architecture in section \ref{sec:model} processes all entities independently, the number of entities can vary between events. See Appendix for the description of events we consider in our dataset and \textbf{\href{https://sites.google.com/site/energyconceptmodels/}{sites.google.com/site/energyconceptmodels}} for video results of our model learning on these events.

\subsection{Understanding Concepts in Multiple Contexts}
\label{sec:experiments-multiple}
An important property of our model is ability to learn from and apply it in both generation and identification contexts. We qualitatively observe that the model performs sensible behavior in both contexts. For example, we considered events with proximity relations \emph{"closest"} and \emph{"farthest"} and found model able to both attend to entities that are closest or furthest to another entity, and to move an entity to be closest or furthest to another entity as shown in figure \ref{fig:prox}. There are multiple admissible solutions which can be generated, as shown by the energy heatmap overlaid. 
We also wish to understand several other properties of this formulation, which we discuss below.
\insertsidefigure{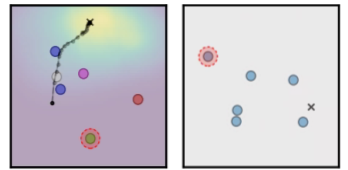}{0.40}{0.5}{Outcomes of generation (left) and identification (right) for the concept of being farthest to cross-shaped entity. Path in left image is the optimization trajectory for the cross entity with the energy heatmap is overlaid.}

\paragraph{Transfer of learning between generation and identification contexts:} When our model trained on both contexts it shares experience between contexts. Knowing how to act out a concept should help in identifying it and vice versa - an effect observed in humans and other animals \cite{acharya2012mirror}. To evaluate the efficacy of transfer, we perform an experiment where we train the energy model only in identification context and test the model's performance in generation context (and conversely and second experiment where we train in generation context and test on identification context). Since it is difficult to quantitatively evaluate generative models which have multiple admissible solutions, we have collected a set of events that only involve the task of moving to an absolute location which have unique answer that allows quantitative evaluation. The results of transfer between contexts on this subset of events are reported in figure \ref{fig:accuracy-transfer}.

We observe that even without explicitly being trained on the appropriate context, the networks perform much better than baseline of two independently-trained networks, though not as effectively as networks that were trained on both contexts. This transfer is advantageous because in many situations demonstrations from only one type of context may be available, which our framework would still be able to integrate.
\insertsidefigure{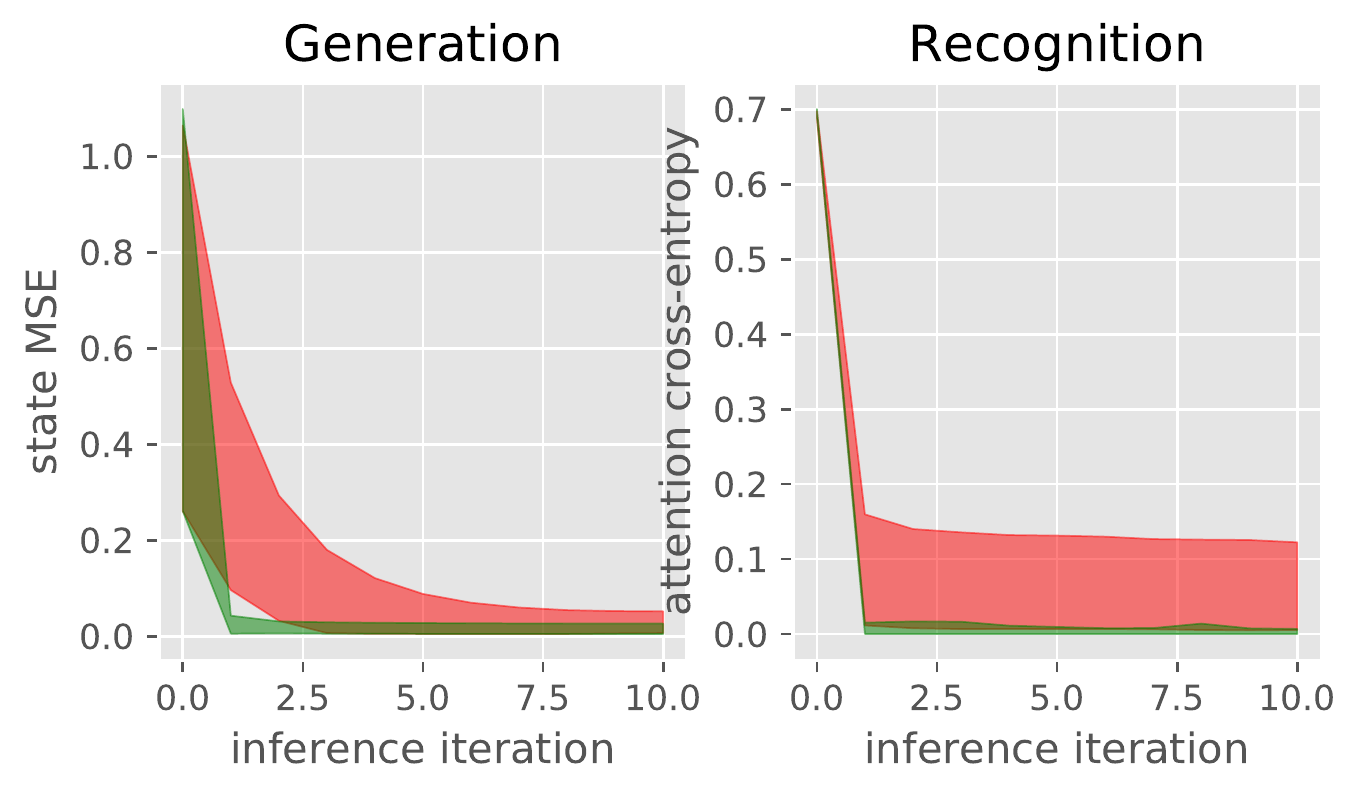}{0.48}{0.5}{Accuracy of transfer between contexts for an absolute position concept. \textbf{Red} is error of the model trained only in one context (generation or identification) evaluated on the opposite context. \textbf{Green} is error of the model trained in both contexts.}

\paragraph{Sharing of concept codes across contexts:}
\insertwrapfigure{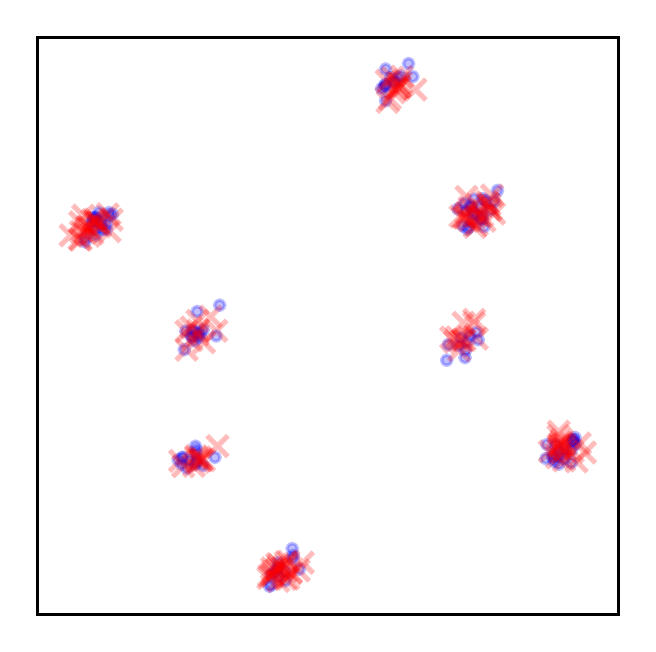}{0.45}{0.30}{Projected concept codes for color events. \textbf{red} are generation codes $\bw_x$ and \textbf{blue} are attention codes $\bw_a$.}
Another property of our model is that codes $\bwx$ and $\bwa$ for identifying concepts are interchangeable and can be shared between generation and identification contexts. For example, either turning an entity red would or identifying all red entities in the scene would ideally use the same concept of \emph{"red"}. We indeed observe that events which involve recognizing entities of a particular color, the codes $\bwa$ match the codes $\bwx$ for setting entities to that color (see Figure \ref{fig:wordvec} for the PCA projection of these codes). We find similar correlation in the other events as well. Thus we see evidence that a concept code is reused across contexts, similarly to how words in a language are used in multiple contexts. This property presents exciting opportunities in applying our model to grounded language understanding.

\subsection{Optimization-Based Inference}
Another important property of our model is that inference processes are based on iterative stochastic optimization dynamics that build up output over time and may involve non-trivial feedback corrections. 
In figure \ref{fig:opt-progress} (left), we show examples of the optimization trajectories for generation of different shape concepts. We see that the multi-step processes consist of a number of non-trivial feedback corrections to achieve the appropriate joint arrangement of entities. On the other hand, a single-step processes must achieve the arrangement through a single very precise step. While this can be adequate for simple shapes such as a line, is it problematic for more complex shapes such as the square.

In optimization trajectories of attention vectors for identification, we observe a mix of outcomes - in some cases attention vector is settled on early in the optimization process, but in other cases optimization involves non-trivial feedback corrections as shown in figure \ref{fig:opt-progress} (top right). In optimization of concept codes, we also observe that desired energy landscape forms only after multiple optimization iterations as from in figure \ref{fig:opt-progress} (bottom right).

\insertfigure{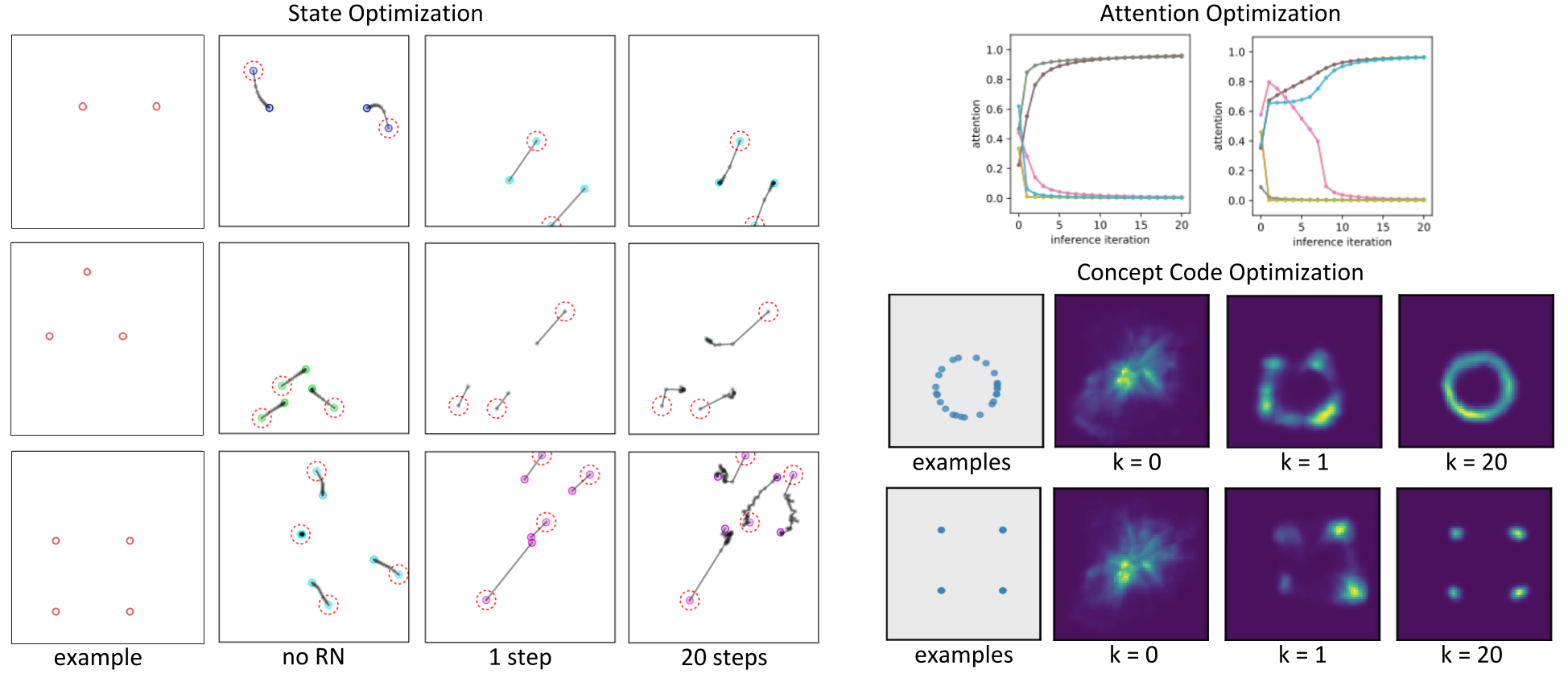}{1.0}{Trajectories of our execution-time inference processes. \textbf{Left:} Generation trajectories for shape concepts (given example shape) trained and executed with 1 optimization step and 20 steps. \textbf{Top Right:} Examples of two attention vector optimization trajectories for identification of a \emph{line} concept. \textbf{Bottom Right:} Examples of energy landscape as $\bw$ optimization progresses.}



\paragraph{Effect of a Relational Architecture} We find that using a relational architecture in our model as in equation \ref{eq:energyfunc} complements the optimization-based inference. We considered an alternative energy function that independently operates over individual entities rather than pair of entities, such as
$
E_\theta(\bx,\ba,\bw) = f_\theta(\sum_{t,i} \sigma(\ba_i) \cdot g_\theta(\bx^t_i, \; \bw), \; \bw)^2
$.
We find that concepts that involve a single entity, such as positioning or color concepts are able to be generated and identified without the use of relation network architecture. However, for concepts that involve coordination of multiple entities such as shape or temporal concepts we observe that not using relation network results in poor samples as shown in \emph{"no RN"} case of figure \ref{fig:opt-progress}.

\paragraph{Effect of Training with $\losskl$ Objective} We wish to understand whether it is necessary to explicitly encourage inference process to produce good samples via $\losskl$ objective in equation (\ref{eq:loss_kl}) as it involves a computationally expensive back-propagation through the optimization procedure. Given enough steps, stochastic gradient Langevin dynamics could in theory generate samples from the energy model's distribution without this explicit objective.

However, in our experiments we observe that training without $\losskl$ objective, the gradient-based inference process of equations in (\ref{eq:sample}) is not able to produce good samples in a small number of steps. Sample negative log-likelihoods are significantly higher when training without $\losskl$ objective. In figure \ref{fig:energy-nokl} we see that while the energy network learns to discriminate between true example events (plotted in green) and sampled and random events (plotted in red and black, respectively) as a result of objective $\loss$. However, the network is unable to produce sample events that match energy of examples. On the other hand, the network trained with both objectives $\loss$ and $\losskl$ is able to generate samples that match the energies of examples while still being able to discriminate between true example and random events.
\insertsidefigure{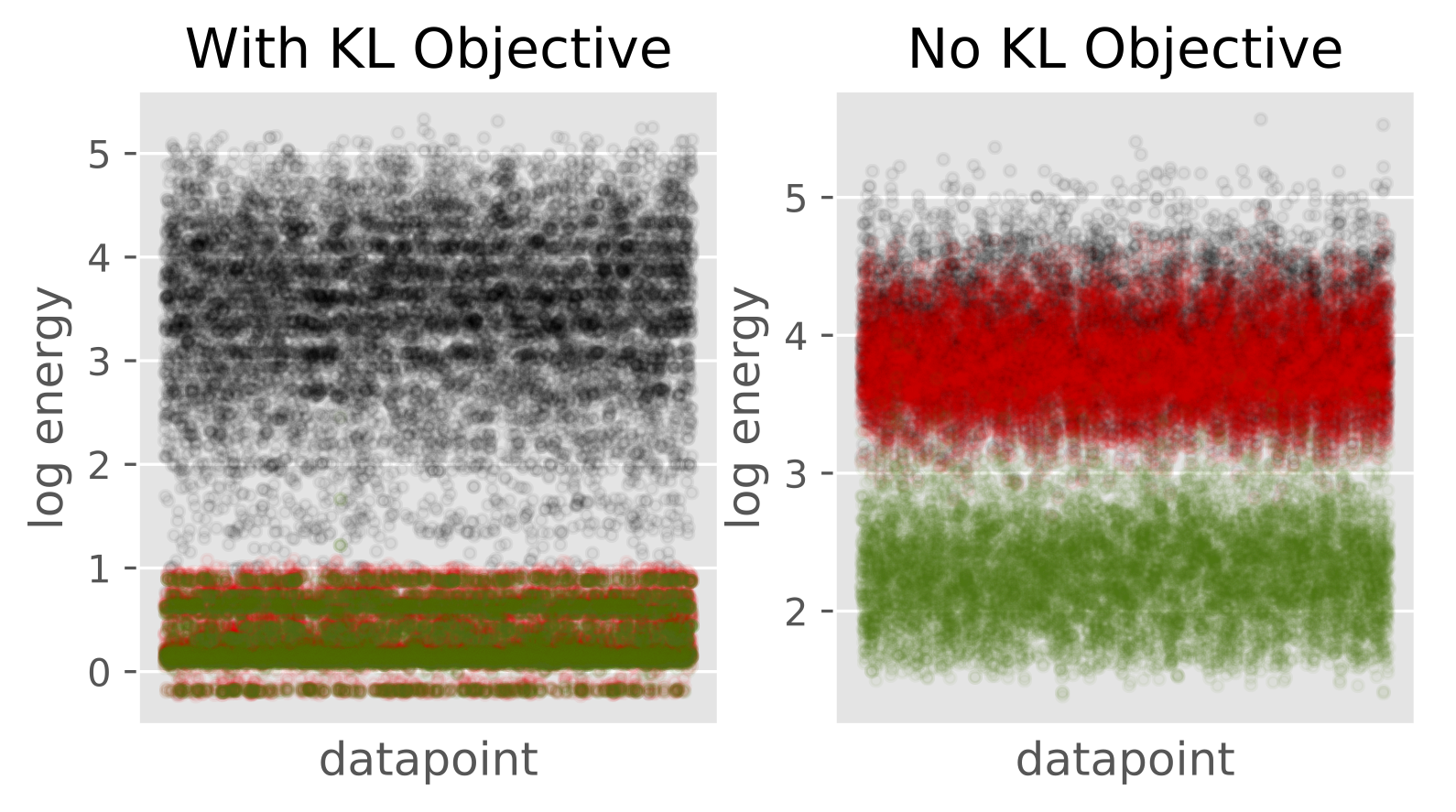}{0.40}{0.5}{Energy values from models trained with and without KL objective. \textbf{Green} and positive example events, \textbf{red} are sample events generated by our run-time inference process, and \textbf{black} are random events from the initial distribution.}

\subsection{Reuse of Concepts between Environments}
When a concept is learned implicitly (as opposed to generated in a feed-forward manner by a generator function or a policy), it allows the possibility to reuse the concept model under different environments and actuation platforms, provided there can be a mapping between the representations of the two environments.

To test generation of concepts in a different environment, we generated similar scenes in a three-dimensional physically-based mujoco environment \cite{todorov2012mujoco} where the actuation mechanism is joint torques of a robotic arm rather than direct changes to environment state. To generate a concept, we used model-predictive control \cite{tassa2012synthesis} as the optimization mechanism and used energy function learned in original environment as a cost for this optimization. Figure \ref{fig:robot} shows results of reusing the concepts to reenact behavior of original demonstration to move into a location between two blue entities. Note that we manually defined a correspondence between representations of two environments and do not claim to tackle automatic transfer of representations - our aim is to show that learned energy functions are robust to being used under different dynamics, actuation mechanism and control algorithm. 
\insertsidefigure{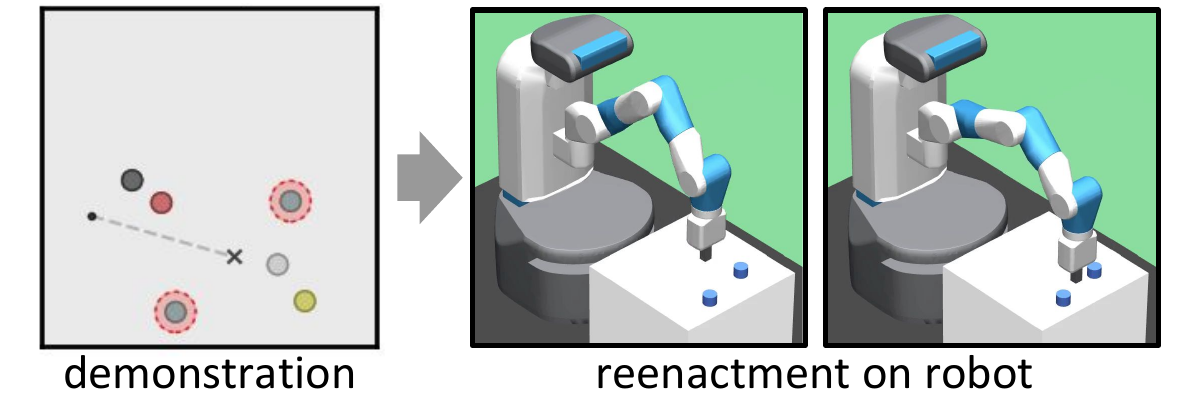}{0.50}{0.4}{Energy function of reaching between blue objects learned from demonstration in 2D particle environment reused in a 3D robot simulator under a novel arrangement.}

\section{Conclusion}
We believe that execution-time optimization plays a crucial role in acquisition and generalization of knowledge, planning and abstract reasoning, and communication. In this preliminary work, we proposed energy-based concept models as basic building blocks over which such optimization procedures can fruitfully operate.
In the current work we used a simple concept structure, but more complex structure with multiple arguments or recursion would be interesting to investigate in the future. It would also be interesting to test compositionality of concepts, which is very suited to our model as compositions corresponds to the summation of the constituent energy functions.





\bibliographystyle{abbrv}
\bibliography{main}

\section*{Appendix}
\label{sec:appendix}
\subsection*{Derivation of Joint Log-Likelihood Approximation}
\label{sec:derivation}
The derivation is similar to guided cost learning \cite{finn2016guided} and cost learning in linearly-solvable MDP \cite{dvijotham2010inverse} formulations. Joint negative log-likelihood of observing tuple $\rbb{\bx^0,\bx^1,\ba}$ is $-\log\cprob{\bx^1,\ba}{\bx^0,\bw}$
\begin{align}
&= -\log\rbb{\cprob{\bx^1}{\ba,\bwx} \cprob{\ba}{\bx^0,\bwa}} \\
&= -\log \frac{\exp\cbb{-E(\bx^1,\ba,\bwx)}}{\int_{\sx} \exp\cbb{-E(\sx,\ba,\bwx)}} - \log\frac{\exp\cbb{-E(\bx^0,\ba,\bwa)}}{\int_{\sa} \exp\cbb{-E(\bx^0,\sa,\bwa)}} \label{eq:logsum}
\end{align}

Consider a more general form of the two individual terms above with non-negative function $f$
\begin{align}
\label{eq:import}
-\log \frac{\exp\cbb{-f(\bx)}}{\int_{\sx} \exp\cbb{-f(\sx)}}
&= f(\bx) + \log\ep{\sx \sim q}{ \frac{\exp\cbb{-f(\sx)}}{q(\sx)}}
\end{align}

The equality follows due to importance sampling under distribution $q$. There are a number of choices for sampling distribution $q$, but a choice that simplifies the above expression and we found to give stable results in practice is $q(X) = \frac{1}{2}\mathbb{I}\sbb{X=\bx} + \frac{1}{2}\mathbb{I}\sbb{X=\sx}$ where $\sx \sim \pi$ and $\pi$ is a distribution that minimizes $\KL{\pi(X)}{\exp\cbb{-f(X)}/Z}$.

In this case, sample-based approximation of equation (\ref{eq:import}) leads to
\begin{align}
f(\bx) + \log\ep{\sx \sim q}{ \frac{\exp\cbb{-f(\sx)}}{q(\sx)}} &\approx f(\bx) + \log\rbb{\exp\cbb{-f(\bx)} + \exp\cbb{-f(\sx)}} \\
&= \log\rbb{1 + \exp\cbb{f(\bx) - f(\sx)}} = \sbb{f(\bx) - f(\sx)}_\softplus
\end{align}
Using the above approximation in equation (\ref{eq:logsum}), gives the desired result $-\log\cprob{\bx^1,\ba}{\bx^0,\bw}$
\begin{align}
\approx& \sbb{E(\bx^1, \ba, \bwx) - E(\sx, \ba, \bwx)}_\softplus + \sbb{E(\bx^0, \ba, \bwa) - E(\bx^0, \sa, \bwa)}_\softplus \\
&\text{where } \sx \sim \csamplex{\bdot}{\ba,\bwx}, \;\; \sa \sim \csamplea{\bdot}{\bx^0,\bwa} \nonumber
\end{align}

\subsection*{Environment Event Descriptions}
We created a dataset containing a variety of events in our environment in order to evaluate learning of a variety of concept in both generation and identification contexts. This dataset is not meant to be an exhaustive list of all possible concepts, but meant to represent a varied sampling of possible interesting concepts. The events involve relations described below, which are either generated or attended to:
\begin{itemize}
\item Changing color of entities or attending to entities of a particular color.
\item Regional placement of entities in a particular spatial area - either a point, horizontal or vertical line, circle, or corners of a square.
\item Placement relations of a one entity either north, south, east, west of another entity or between two entities.
\item Shape relations between entities of either joining together, or forming a line, triangle, or square shapes.
\item Proximity relations bring attention to the entity closest or furthest to another entity or to bring that entity to be closest or furthest to the attended entity.
\item Quantity relations bring attention to any one, two, three, or more than three entities.
\item Temporal relations bring an one entity to another entity only after the other entity starts moving.
\end{itemize}
See \textbf{\href{https://sites.google.com/site/energyconceptmodels/}{sites.google.com/site/energyconceptmodels}} for video results of our model learning on these events.

\subsection*{Experiment and Training Details}
In all our experiments, $f$ and $g$ in the relation network in section \ref{sec:model} are multi-layer neural networks with two hidden layers and 128 hidden units each. We use $5$ demonstration examples in $X^\demo$ and $X^\train$ for each concept and use a batch size of 1024. We use $K=10$ gradient descent steps for concept inference and sampling. We trained our experiments for 10000 timesteps and used Adam \cite{kingma2014adam} optimizer with learning rate $10^{-3}$.

\subsection*{Additional Results}
We provide quantitative results for the transfer of learning between generation and identification contexts described in Section \ref{sec:experiments-multiple}.
\inserttable{reward}{|c|c|c|}{
\hline
\textbf{Condition} & \textbf{Generation Error} & \textbf{Attention Error} \\
\hline
Untrained Network & 0.621 & 0.698 \\
\hline
Trained on Generation & 0.006 & 0.061 \\
\hline
Trained on Identification & 0.016 & 0.001 \\
\hline
Trained on Both & 0.007 & 0.001 \\
\hline}{Training and test physical reward for setting with and without communication.}

\subsection*{Algorithm Details}
For completeness, we provide the algorithm of our method below.

\begin{algorithm}[H]
 \SetAlgoLined
  \begin{algorithmic}
    \STATE Initialize energy model parameters $\theta$  
    \FOR{events $X^\train$ and $X^\demo$ sampled from the same concept}
      \STATE Randomly sample event $\rbb{\bx^0, \bx^1, \ba}$ from $X^\demo$
      \STATE Initialize $\bwx$ and $\bwa$ from unit Gaussian
      \FOR{sampling iteration $k = 1$ to $K$}
        \STATE Update samples $\bwx$ and $\bwa$ via stochastic gradient Langevin dynamics step:
        \begin{align*}
            \bwx(\theta) &\gets \bwx(\theta) + \frac{\alpha}{2} \nabla_\bw E_\theta(\bx^0, \ba, \bwx(\theta)) + \omega^k \\
            \bwa(\theta) &\gets \bwa(\theta) + \frac{\alpha}{2} \nabla_\bw E_\theta(\bx^1, \ba, \bwa(\theta)) + \omega^k  
        \end{align*}
      \ENDFOR
      \STATE Randomly sample event $\rbb{\bx^0, \bx^1, \ba}$ from $X^\train$

      \STATE Initialize $\tilde{\bx}$ to $\bx^0$ and initialize $\tilde{\ba}$ from unit Gaussian
      \FOR{sampling iteration $k = 1$ to $K$}
        \STATE Update samples $\tilde{\bx}$ and $\tilde{\ba}$ via stochastic gradient Langevin dynamics step:
        \begin{align*}
            \tilde{\bx}(\theta) &\gets \tilde{\bx}(\theta) + \frac{\alpha}{2} \nabla_\bx E_\theta(\tilde{\bx}(\theta), \ba, \bw_x(\theta)) + \omega^k \\
            \tilde{\ba}(\theta) &\gets \tilde{\ba}(\theta) + \frac{\alpha}{2} \nabla_\ba E_\theta(\bx^0, \tilde{\ba}(\theta), \bw_a(\theta)) + \omega^k    
        \end{align*}
      \ENDFOR
    
      \STATE Set $\bar{\bx}$ be the result of applying gradient stopping operator on $\sx$ (and similarly for $\bar{\ba}$, $\bar{\bw}$ and $\bar{E}$) 
      \STATE Formulate two losses operating over $p$ and $\pi$ holding other function fixed:
      \begin{align*}
        \loss(\theta) &= \sbb{E_\theta(\bx^1, \ba, \bwx(\theta)) - E_\theta(\bar{\bx}, \ba, \bwx(\theta))}_\softplus + \sbb{E_\theta(\bx^0, \ba, \bwa(\theta)) - E_\theta(\bx^0, \bar{\ba}, \bwa(\theta))}_\softplus \\
        \losskl(\theta) &= \bar{E}(\sx(\theta), \ba, \bar{\bw}_x) + \bar{E}(\bx^0, \sa(\theta), \bar{\bw}_a)          
      \end{align*}
      \STATE Update $\theta$ based on gradient $\nabla_\theta(\loss(\theta) + \losskl(\theta))$ via Adam optimizer
    \ENDFOR
  \end{algorithmic}
 \caption{Energy-based model learning from demonstration events}
\end{algorithm}

\end{document}